\documentclass[conference]{IEEEtran}
\IEEEoverridecommandlockouts
\usepackage{cite}
\usepackage{amsmath,amssymb,amsfonts}
\usepackage{algorithmic}
\usepackage{graphicx}
\usepackage{textcomp}
\usepackage{xcolor}
\usepackage{enumitem} 
\usepackage{stfloats}
\usepackage{placeins}
\usepackage{todonotes}
\usepackage{hyperref}
\usepackage{tabularx}
\usepackage{multicol}
\usepackage{array,multirow}
\usepackage{pgfplots}
\usepackage{pgfplotstable}
\usepackage{xcolor}
\def\BibTeX{{\rm B\kern-.05em{\sc i\kern-.025em b}\kern-.08em
    T\kern-.1667em\lower.7ex\hbox{E}\kern-.125emX}}
\begin{document}

\title{Fair-FLIP: Fair Deepfake Detection with Fairness-Oriented Final Layer Input Prioritising

\thanks{Tomasz Szandała and Fatima Ezzeddine are supported by the Swiss Government Excellence Scholarships. This work is partially supported by the Swiss National Science Foundation (grant number CRSII5\_209250).\\Corresponding author: Tomasz.Szandala@supsi.ch.}
}

\author{\IEEEauthorblockN{Tomasz Szandala\textsuperscript{†}, Fatima Ezzeddine\textsuperscript{†,*}, Natalia Rusin\textsuperscript{‡}, Silvia Giordano\textsuperscript{†} and Omran Ayoub\textsuperscript{†}}
\IEEEauthorblockA{
\textit{\textsuperscript{†} University of Applied Sciences and Arts of Southern Switzerland, Lugano, Switzerland}\\
\textit{\textsuperscript{*} Università della Svizzera italiana, Lugano, Switzerland}\\
\textit{\textsuperscript{‡} Independent Researcher}
}
}

\maketitle

\begin{abstract}
Artificial Intelligence-generated content has become increasingly popular, yet its malicious use, particularly the deepfakes, poses a serious threat to public trust and discourse. While deepfake detection methods achieve high predictive performance, they often exhibit biases across demographic attributes such as ethnicity and gender. In this work, we tackle the challenge of fair deepfake detection, aiming to mitigate these biases while maintaining robust detection capabilities. To this end, we propose a novel post-processing approach, referred to as Fairness-Oriented Final Layer Input Prioritising (Fair-FLIP), that reweights a trained model's final-layer inputs to reduce subgroup disparities, prioritising those with low variability while demoting highly variable ones. Experimental results comparing Fair-FLIP to both the baseline (without fairness-oriented de-biasing) and state-of-the-art approaches show that Fair-FLIP can enhance fairness metrics by up to 30\% while maintaining baseline accuracy, with only a negligible reduction of 0.25\%. 

Code is available on \emph{Github}:
\url{https://github.com/szandala/fair-deepfake-detection-toolbox}
\end{abstract}


\section{Introduction}
The diffusion of Artificial Intelligence (AI)-generated content has accelerated in recent years, driven by the increasing sophistication of generative algorithms~\cite{cao2025survey}. Advances in deep learning, particularly in natural language processing and computer vision, have enabled AI systems to produce convincingly realistic images, videos, audio, or text with minimal human oversight~\cite{cao2025survey,rana2022deepfake,dong2022explaining}.
Despite the benefits in domains such as education and entertainment, these technologies remain susceptible to misuse, posing significant societal implications~\cite{chesney2019deep,salvi2024conversational}. 

Among the most critical and potentially harmful manifestations of AI-generated content is the \emph{deepfake}. As defined by Chesney and Citron~\cite{chesney2019deep}, deepfake is a synthetic or manipulated digital media, e.g. video or audio, convincingly simulating real individuals or events. Deepfakes pose a serious threat to the modern digital society by enabling the rapid creation and distribution of persuasive misinformation, often with malicious intent~\cite{guarnera2020deepfake,narvali2023cyberbullying}. This danger has led to alarming events usually targeting high-profile individuals but also ordinary people; for instance, manipulated videos of the former U.S.\ President Barack Obama delivering fictional remarks or privacy-violating hoaxes of local high-school students~\cite{davisson2024eros,narvali2023cyberbullying}. 

Beyond reputational and psychological harm, deepfakes fuel distrust in public discourse, undermine the credibility of genuine news, and thus may destabilise political processes~\cite{swiss_democracy2024}. A recent report from Lausanne researchers shows that fake news supported by AI-generated content tends to be 82\% more persuasive than traditional misinformation, thus posing a threat to direct democracies like the Swiss one~\cite{swiss_democracy2024,salvi2024conversational}.

Consequently, research into deepfake detection has expanded significantly, driven by the need to protect individuals and institutions from harmful exploitation~\cite{rana2022deepfake,malik2022deepfake,dong2022explaining,raza2022novel,guarnera2020deepfake,agarwal2021deepfake,aghasanli2023interpretable}. Although these efforts have significantly advanced detection accuracy, the question of \emph{fairness} in deepfake detection remains relatively under-explored. Fairness can be defined as the model's behaviour, ensuring it is not systematically biased to misclassify individuals based on attributes such as ethnicity or physical features unrelated to the classification task~\cite{dwork2012fairness}. It is essential to emphasise that unfair detection systems risk provide unequal protection, especially to less privileged groups who may lack the power or resources to contest false content. This can lead to the erosion of trust in such systems~\cite{matli2024extending}. 

Recent studies have begun exploring \emph{bias mitigation} in deepfake detection; however, many existing approaches (discussed more later on) rely on retraining with protected attributes, threshold adjustments, or specialised sub-models, limiting generalisation and complicating deployment.
In response, our work aims to bridge this gap by proposing a targeted de-biasing strategy that complements existing bias mitigation methods. 

In this work, we propose \emph{Fair-FLIP} (\emph{Fairness-Oriented Final Layer Input Prioritising}), a novel post-processing method designed to mitigate subgroup disparities by reweighting the final-layer features of an already-trained deepfake detector. Fair-FLIP subtly alters the importance of each feature without retraining, prioritising those that exhibit low subgroup variability while reducing the impact of highly variable features across protected groups. The proposed approach is applicable to any neural network model, requires no demographic knowledge at inference, and avoids significant architectural or training changes, thereby minimising the development and computational costs. In comparison with state-of-the-art techniques, our proposed approach demonstrates superior results by preserving baseline accuracy within 0.25\% of the original while delivering an average 10\% improvement in fairness metrics.

The rest of the paper is organised as follows. Sec. \ref{sec:metrics} overviews fairness metrics and bias mitigation strategies. Sec. \ref{sec:sota} discusses related work. Sec. \ref{fairflip} introduces our proposed approach. Sec. \ref{results} details the experimental setup and discusses numerical results. Finally, Sec. \ref{sec:discussion} presents further analysis and discusses limitations and Sec. \ref{sec:conclusions} concludes the paper. 

\vspace{-0.2cm}
\section{Fairness Metrics and Bias Mitigation}\label{sec:metrics}


\subsection{Fairness Metrics}\label{sec:fair-metr}
The literature identifies numerous metrics to quantify the fairness of the model, primarily focusing on performance comparisons across protected attributes such as, e.g., gender, ethnicity, or physical features ~\cite{barocas2019fairness}. In this section, we introduce four fairness metrics that broadly capture key aspects of evaluation.
We consider True Positive Parity (TPP), False Positive Parity (FPP), Positive Predictive Value (PPV), and Negative Predictive Value (NPV) as our fairness metrics. They collectively provide a comprehensive view of model performance across different protected categories while highlighting error distribution.

Let $a,b,c \in A$ represent different protected attribute values~(e.g. ethnicity white, black, asian, etc.), and let $TP_a, TN_a, FP_a, FN_a$ be each attribute's respective true/false positives/negatives.

\textbf{Equality of Odds}. Equality of Odds metric requires that \emph{True Positive Parity} (TPP) and \emph{False Positive Parity} (FPP) rates be consistent across groups~\cite{hardt2016equality}. TPP~(eq.~\ref{eq:tpp}), also referred to as Equality of Opportunity, ensures that individuals in the positive class have the same chance of correct classification across groups. On the contrary, FPP~(eq.~\ref{eq:fpp_single}) checks that the rate of false positives is not excessive for certain groups. 
In other words, skewed TPP means certain groups are recognised more accurately than others, while skewed FPP means some groups are more often mislabelled as positive.

\begingroup
\setlength{\multicolsep}{1.2ex}
\setlength{\parskip}{0pt}      
\setlength{\abovedisplayskip}{-1ex}  
\begin{multicols}{2}
\begin{equation}
TPP_a = \frac{TP_a}{TP_a + FN_a}
\label{eq:tpp}
\end{equation}

\begin{equation}
FPP_a = \frac{FP_a}{FP_a + TN_a} 
\label{eq:fpp_single}
\end{equation}
\end{multicols}

\textbf{Predictive Value Parity}. Predictive Value Parity checks whether the fraction of correct predictions among all positive or negative predictions is equal across groups~\cite{chouldechova2017fair}. \emph{Positive Predictive Value} (PPV) measures how reliably a positive prediction corresponds to a true positive~(eq.~\ref{eq:ppv_single}), while \emph{Negative Predictive Value} (NPV) measures how reliably a negative prediction corresponds to a true negative~(eq.~\ref{eq:npv_single}). A skew in PPV implies that positive predictions are more accurate for some groups, while a skew in NPV indicates that negative predictions are more reliable for specific groups.

\begin{multicols}{2}
\begin{equation}
PPV_a = \frac{TP_a}{TP_a + FP_a}
\label{eq:ppv_single}
\end{equation}

\begin{equation}
NPV_a = \frac{TN_a}{TN_a + FN_a} 
\label{eq:npv_single}
\end{equation}
\end{multicols}
\endgroup

Any imbalance in parity, the ratio of lowest to highest value~(eq.~\ref{eq:metric_ratio}), across any of these metrics signals bias in prediction. In our work, we aim to bring all fairness metrics as close as possible to $1.0$ while maintaining accuracy close to the original.

\vspace{-0.5ex}
\begin{equation}
metric\_parity = \frac{\min(metric_a, metric_b, metric_c,\dots)}{\max(metric_a, metric_b, metric_c,\dots)}
\label{eq:metric_ratio}
\end{equation}

\subsection{Model De-Biasing Techniques}
\label{sec:debiasing}
Researchers primarily employ methods for model de-biasing at three phases in a machine learning pipeline, namely, \emph{(1) pre-processing data}, \emph{(2) in-processing model training}, and \emph{(3) post-processing model's outputs}. Each method has inherent trade-offs regarding effectiveness, complexity and computational resources.

\subsubsection{Pre-Processing Techniques}
Pre-processing methods focus on refining input data to reduce unfair patterns prior to model training. Traditional approaches include:
\begin{itemize}[noitemsep,topsep=0pt,leftmargin=*]
    \item \textbf{Reweighting and Resampling}~\cite{kamiran2012classifying}: Assigning different importance weights or sampling rates to protected and non-protected groups to ensure balanced representation.
    \item \textbf{Latent Feature Transformation}~\cite{zemel2013learning}: Transforming the feature space to obfuscate sensitive attributes whilst preserving predictive signals.
    \item \textbf{FairTest}~\cite{tramer2017fairtest}: introduces a framework for detecting unfair or discriminatory treatment in data-driven systems by broadening group fairness definitions, providing guidelines for applying fairness principles at the stage of model and dataset preparation.
\end{itemize}

The main drawbacks of pre-processing techniques are that they may inadvertently distort key characteristics or oversimplify complex correlations and fail to fully eliminate bias~\cite{szandala2024discriminating}. 

\subsubsection{In-Processing Techniques}
In-processing involves integrating fairness constraints, or penalties, in the model training process (e.g., in the loss function of the employed model). This approach compels the model to account for fairness requirements during optimisation explicitly. This may lead to a reduction in overall accuracy when fairness and performance objectives conflict~\cite{tizpaz2022fairness}. Moreover, their application includes relatively high computational overhead during training, extending the model's training time. These methods often employ:
\begin{itemize}[noitemsep,topsep=0pt,leftmargin=*]
    \item \textbf{Fairness-Constrained Optimisation}~\cite{zafar2017fairness,cotter2019training}: Adding fairness constraints (e.g., demographic parity) to the loss function.
    \item \textbf{Adversarial Fairness}~\cite{edwards2015censoring,zhang2018mitigating}: Training a predictor simultaneously attempting to confuse an adversary predicting sensitive attributes.
\end{itemize}


\subsubsection{Post-Processing Techniques}
Post-processing alters model predictions after training, offering a relatively simpler to enforce fairness with respect to the above-mentioned techniques. Example techniques include:
\begin{itemize}[noitemsep,topsep=0pt,leftmargin=*]
    \item \textbf{Group-Specific Thresholding}~\cite{hardt2016equality,zhao2021calibrating}: Adjusting decision thresholds per demographic group to achieve targets like equalised odds. Alternatively, we can modify the threshold globally when we do not want to provide a protected attribute during inference.
    \item \textbf{Reject-Option Classifiers}~\cite{kamiran2012classifying}: Overriding uncertain predictions based on sensitive attributes.
\end{itemize}

A key limitation of these techniques is that they may introduce inconsistencies, compromising the model's reliability. Additionally, because these fairness adjustments are imposed artificially rather than naturally learned during training, they can obscure the model's reasoning process, making explainability more challenging~\cite{siniosoglou2023post}.

\section{Related Work}\label{sec:sota}

Several studies have focused on detecting deepfake across images, audio, and video modalities \cite{rana2022deepfake,malik2022deepfake,dong2022explaining,raza2022novel,guarnera2020deepfake,agarwal2021deepfake,aghasanli2023interpretable}, leveraging statistical techniques \cite{koopman2018detection,baar2012camera,welch1947generalization,agarwal2019limits}, ML-based \cite{zhang2019detecting,sahla2021detection,yang2019exposing} or deep learning-based methods \cite{feng2020deep,afchar2018mesonet,nirkin2020deepfake,qi2020deeprhythm,hernandez2020deepfakeson,du2020towards}.


For example, Matern et al.~\cite{matern2019exploiting} examine subtle facial-region manipulations, such as changes in eye shading or added accessories that can be easily overlooked. In \cite{sahla2021detection} authors propose a lightweight multi-layer perceptron to detect visual artifacts in the face region with minimal computational overhead. A more advanced strategy uses a Generative Adversarial Network (GAN) simulator that synthesises collective GAN-generated image artefacts as input features for a classifier, thereby enhancing the capacity to distinguish deepfakes~\cite{zhang2019detecting}. In~\cite{raza2022novel}, authors propose a multi-model convolutional neural network (CNN) to further boost artefact detection accuracy. Other notable studies include frequency-domain analysis to distinguish real from manipulated images~\cite{agarwal2021deepfake}, and an analyser of tensor-based representation of compressed and resized images enriched by discrete cosine transform features~\cite{concas2022tensor}. More recent works explore refined model architectures, such as fine-tuned Vision Transformers (ViTs) integrated with support vector machines~\cite{aghasanli2023interpretable} or CNNs~\cite{arshed2023unmasking}, to effectively detect deepfake images generated by Diffusion Models. 

Fair deepfake detection, including the assessment of fairness and bias, has increasingly drawn attention recently. In~\cite{pu2022fairness}, the authors highlight a bias factor of approximately 4\% in favour of female subjects when evaluating MesoInception-4. Similar biases are prevalent in other areas of computer vision and facial recognition~\cite{mehrabi2021survey}, where specific subgroups—such as non-Caucasian individuals or those wearing glasses—experience disproportionately high error rates~\cite{trinh2021examination,stroebel2023systematic}. Moreover, studies on XceptionNet~\cite{agarwal2024deepfake} reveal performance drops exceeding 30\% for certain ethnicities and genders.

In terms of proposed de-biasing methods, Ezeakunne et al.~\cite{ezeakunne2024deepfake} proposed a pre-processing technique to generate synthetic images by applying transformations such as scaling and rotation to balance under-represented demographics. Ju et al.~\cite{ju2024improving} proposed in-processing demographic-agnostic and demographic-aware methods that modify the training objective by incorporating fairness metrics, ensuring a more uniform treatment of subpopulations. In \cite{lin2024preserving}, authors propose a framework that detaches demographic and forgery features, albeit at the cost of higher complexity. Despite their effectiveness, they significantly increase training overhead~\cite{han2023ffb} and require detailed model modifications, which are not always feasible~\cite{zafar2017fairness}. In~\cite{liu2024thinking}, authors propose a post-processing technique, namely, Bias Pruning with Fair Activations (BPFA), which prunes network weights that exhibit demographic-related biases. While this approach can yield fairer results, it often degrades performance, and the procedure can be time-consuming or unstable. 

Similar to previous works, our study focuses on mitigating bias in deepfake detection techniques. However, we take a post-processing approach that requires minimal modifications to the trained model, thereby preserving its predictive performance while ensuring it remains lightweight. Our approach operates on the hypothesis that activations exhibiting high variance across protected attributes tend to encode ethnicity-specific features, while those with lower variance capture more general, ethnicity-independent characteristics. Leveraging this insight, we prioritise final-layer inputs accordingly. The most closely related work to ours is BPFA, which mitigates bias by entirely removing certain weights. In contrast, our proposed approach modifies weights selectively, ensuring that essential features are not lost. For a comprehensive evaluation, we compare the performance of our method against Fair-FLIP in Section \ref{sec:results_comparison}.

\section{Problem Formulation and Methodology}\label{fairflip}
 
\subsection{Problem Formulation}
The problem can be formally stated as follows. Given a dataset containing real and fake images across various ethnicities, our goal is to develop a technique that improves the fairness metrics outlined in Sec. \ref{sec:metrics}, while preserving high predictive performance, measured by accuracy and F1-score. Ethnicity is treated as a protected attribute, and fairness metrics are computed accordingly. Note that the de-biasing process must not require the model to be aware of protected attributes during training or inference, nor should it demand substantial architectural alterations or computational resources to be applied. Therefore, our final objective is a lightweight, post-processing method that enhances fairness by countering demographic imbalances without compromising the usability and performance of the original deepfake detector.

\subsection{Fair-FLIP}
\begin{figure*}[t]
    \centering
    \includegraphics[width=1.03\linewidth]{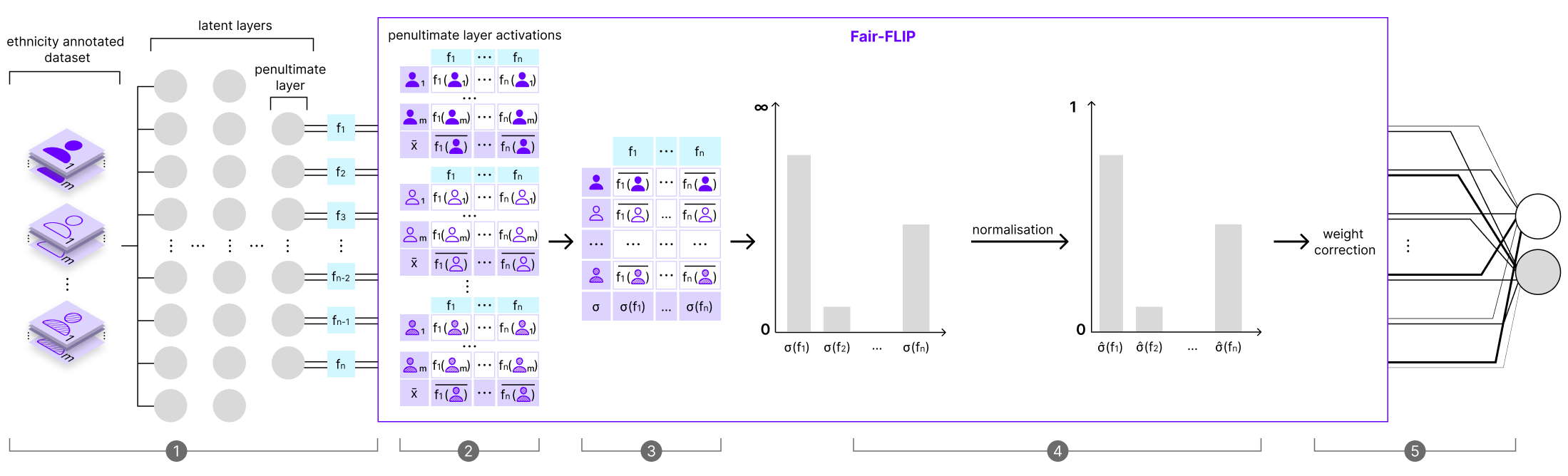}
    \caption{Schematic representation of Fair-FLIP's methodology.}
    \label{fig:f-lip}
    \vspace{-0.3cm}
\end{figure*}

Fair-FLIP is a post-processing technique designed to mitigate subgroup disparities by intervening in a trained deepfake detection model. Its core principle is to down-weight activations of the second-to-last layer, often called \emph{features} on which the classification bases, that exhibit high variability across protected subgroups while promoting those with lower variability. This is based on the hypothesis that activations with relatively high variance across protected attributes are biased to ethnicity-specific features, whereas those with lower variance capture more general, non-ethnicity-dependent characteristics. We chose to apply de-biasing exclusively to the final layer because deeper layers typically encode more abstract features relevant to classification~\cite{zeiler2014visualizing}, making it possible to reduce bias without extensive altering of multiple earlier layers effectively. By doing so, we aim to mitigate biases present across ethnicities. Overall, Fair-FLIP does not require retraining and is equally applicable to any neural network base architectures, such as ViTs, CNNs, or even a discriminator within a GAN—making it a highly versatile solution.

Figure~\ref{fig:f-lip} shows a schematic representation of the working of Fair-FLIP, which we describe in the following steps: 

\begin{enumerate}[leftmargin=*] 
\item \textbf{Capture penultimate-layer activations}: For each image in the ethnicity-annotated dataset, record the activation vector from second-to-last layer of the deepfake detection model. 

\item \textbf{Group activations by ethnicity}: Group the activation vectors according to their associated ethnic subgroup and calculate the mean value of activations for each subgroup.  

\item \textbf{Measure between-group variability}: For each activation $f_i$, compute the standard deviation of its subgroup-specific mean activations, $\sigma_i(f_i)=std(\bar{f_i})$). 

\item \textbf{Normalise variability}: In order to provide comparable values for altering weights, we have to transform each standard deviation into a value within the range $[0,1]$. We achieve this by normalisation: $\hat{\sigma}(f_i)= \frac{\sigma_i - min(\sigma)}{max(\sigma)- min(\sigma)}$.

\item \textbf{Adjust final-layer weights}: Use the normalised standard deviation to modify the original feature weights $w_i$ according to Eq.~\ref{eq:fflip}, where $\alpha$ is a term controlling how strongly each feature is promoted or demoted. 
\end{enumerate}

\vspace{-1ex}
\begin{equation} 
w'_{i} = w_i \times \bigl(1 + \alpha - \hat{\sigma}(f_i)\bigr), 
\label{eq:fflip} 
\end{equation} 

As a result, the weights of the final layer will be modified so that features heavily influenced by ethnicity are suppressed, while more general, reliable signals remain central to the detector's decision-making.

\subsection{Benchmark Approaches}
We compare the performance of Fair-FLIP to a baseline model and state-of-the-art bias mitigation techniques. 

\textbf{Baseline}. The baseline model is a ViT-based classifier trained to label each image as \emph{authentic} or \emph{deepfake}, with no regard to fairness. It represents the model with the highest predictive performance. 

\textbf{Pre-processing.} To reduce data-level bias, we perform pre-processing by undersampling overrepresented ethnic subgroups. This preserves the original model architecture and training procedure as it does not add or alter any hyper-parameters. However, since some examples from overrepresented subgroups are discarded, potentially informative data is lost, harming accuracy. We opt for simple undersampling over more sophisticated methods (e.g. Tomek links, SMOTE, ADASYN) to avoid artefacts and simplify the pipeline.

\textbf{In-Processing.} Building on Zafar \emph{et al.}~\cite{zafar2017fairness} in-processing strategy, we embed fairness penalties directly in the loss function. Specifically, we augment the training objective~(eq. \ref{eq:fair_loss}), encouraging balanced performance across subgroups. Increasing $\lambda$ intensifies the fairness objective at the cost of diminishing accuracy.
\vspace{-1ex}
\begin{equation}
    \text{total\_loss} = \text{loss} + \lambda \,(1 - \text{TPP})^2 \,(1 - \text{FPP})^2
    \label{eq:fair_loss}
\end{equation}

\textbf{Threshold-based Post-Processing.} In threshold-based post-processing, we adjust the decision threshold~(eq. \ref{eq:fof}) to maximise a fairness objective~\cite{zhao2021calibrating}.
\vspace{-1ex}
\begin{equation}
    \text{TPP}^2 \times \text{FPP}^2 
    \times \Bigl(1-\max(\text{FPP}_a, \text{FPP}_b, \dots)\Bigr)^2
    \times \text{PPV}^2 \times \text{NPV}^2
    \label{eq:fof}
\end{equation}

This approach seeks to balance the True/False Positive Parities (TPP, FPP) and predictive values (PPV, NPV) without retraining. A global threshold from $0.0$ to $1.0$ is explored in increments of $0.001$ to identify the configuration that maximises fairness. While this technique optimises fairness, it may come at the cost of a significant decline in accuracy.

\textbf{Bias Pruning with Fair Activations.} We include another post-processing strategy, namely, \emph{Bias Pruning with Fair Activations (BPFA)}~\cite{liu2024thinking}, which removes specific weights deemed responsible for demographic-related biases in a post-processing manner. This approach does not alter the data or training procedure but refines network parameters to mitigate biased representations.

\section{Numerical Results}\label{results}

\subsection{Experimental Settings}
We used a Kaggle dataset of 190{,}335 face images, evenly split between manipulated (95{,}134) and authentic (95{,}201)~\cite{dataset}. Each 256$\times$256 JPEG was assigned an ethnicity label. Ethnicity annotations (White, Black, Asian, Indian, Latino, Middle Eastern) were obtained using a 97\% accurate facial attribute model (HyperExtended LightFace~\cite{serengil2021hyperextended}) and later verified by manual inspection.

\begin{figure}[t]
\vspace{-0.5cm}
    \centering
\includegraphics[width=\linewidth,keepaspectratio]{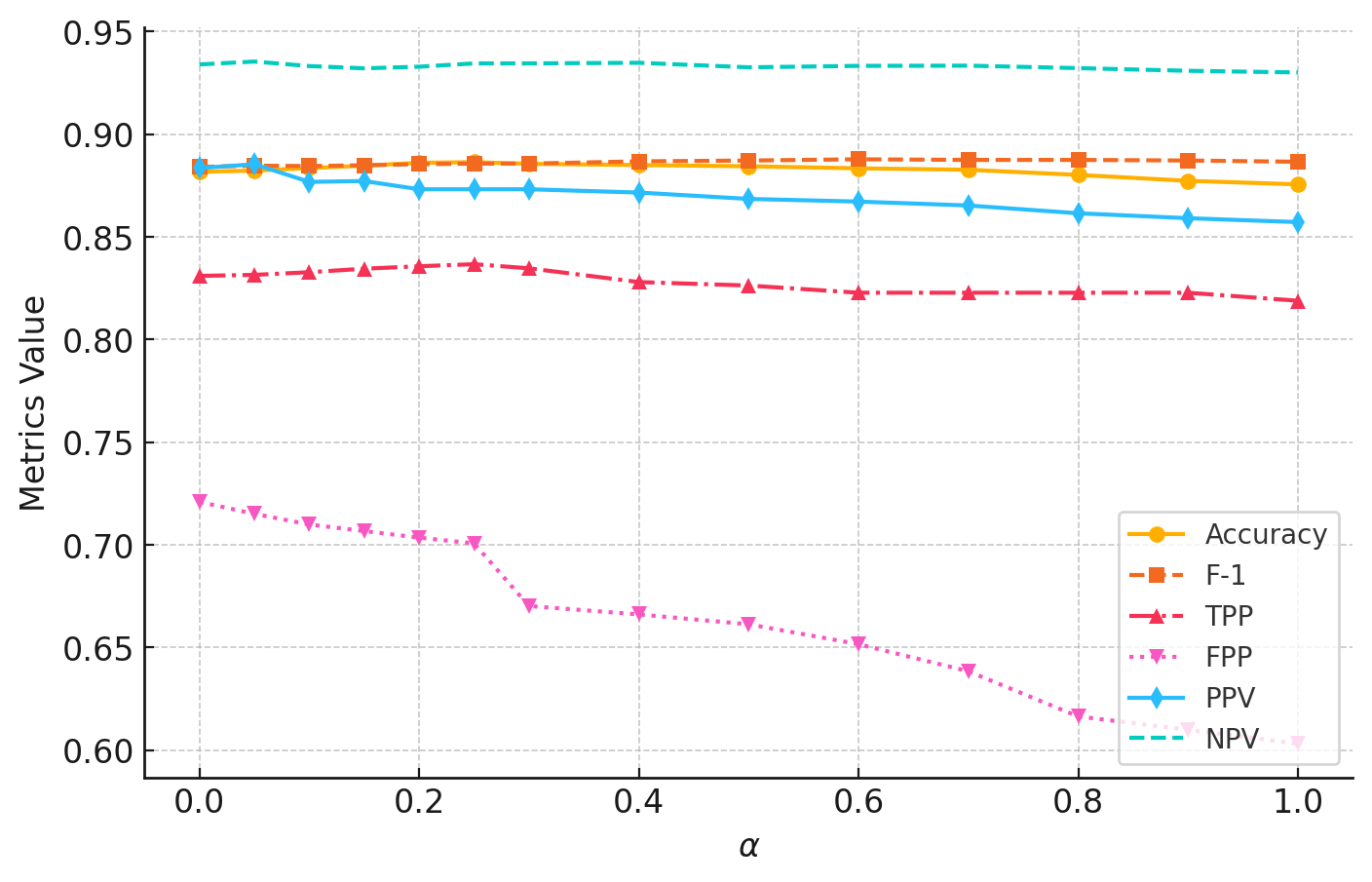}
    \caption{Sensitivity analysis of trade-offs between fairness metrics and accuracy as \texorpdfstring{$\alpha$}{alpha} changes.}
    \label{fig:alpha}
    
\end{figure}

Our baseline model for experiments is a Vision Transformer~\cite{arshed2023unmasking}~(\texttt{google/vit-base-patch16-224-in21k}, pre-trained on \texttt{ImageNet-21k}) trained for 15 epochs with standard cross-entropy loss. We used Adam~(LR=0.001, $\beta=(0.9,0.999)$) in \texttt{PyTorch 1.13.1+cu116}, halving the learning rate every five epochs, with a batch size of 64 and no data augmentation. 

As evaluation metrics, we consider accuracy and F1-score as metrics for predictive performance and TPP, FPP, PPV and NPV as fairness metrics.

\subsection{Determining the value of  \texorpdfstring{$\alpha$}{alpha}}

To determine the value of $\alpha$ parameter of Fair-FLIP for the subsequent experiments, we performed a sensitivity analysis varying the value of $\alpha$. 

Figure~\ref{fig:alpha} reports the accuracy and fairness metrics achieved for values of $\alpha$ ranging between 0.0 and 1.0. Results show that $\alpha$ may have a crucial impact on NPV while not heavily impacting other metrics. For our experiments, we select $\alpha=0.25$, which achieves the best balance between accuracy and fairness metrics, in our opinion. 

\begin{table*}[bht!]
\caption{Table summarising different metrics for model's performance during five-fold cross-validation}
\label{tab:all_approaches}
\begin{tabular}{r|rrrrrrr}
\hline
                                                    & \textbf{Fold}       & \textbf{Baseline}        & \textbf{Pre-Process}     & \textbf{In-Process}       & \textbf{Post-Process (Threshold)} & \textbf{Pruning (BPFA)}  & \textbf{Fair-FLIP (ours)} \\ \hline
\parbox[t]{2mm}{\multirow{6}{*}{\rotatebox[origin=c]{90}{\textbf{Accuracy}}}}                  & 1                   & 0.8865                   & 0.7561                   & 0.8681                    & 0.8772                            & 0.8637                   & 0.8856                    \\
                                                    & 2                   & 0.8880                   & 0.7902                   & 0.8532                    & 0.8391                            & 0.8602                   & 0.8846                    \\
                                                    & 3                   & 0.8715                   & 0.7537                   & 0.8496                    & 0.8226                            & 0.8437                   & 0.8691                    \\
                                                    & 4                   & 0.8520                   & 0.7342                   & 0.7921                    & 0.8031                            & 0.8332                   & 0.8486                    \\
                                                    & 5                   & 0.8677                   & 0.7899                   & 0.8278                    & 0.8188                            & 0.8429                   & 0.8663                    \\
                                                    & \textbf{Avg. (Std)} & \textbf{0.8731 (0.0148)} & \textbf{0.7648 (0.0245)} & \textbf{0.8382 (0.0295)}  & \textbf{0.8322 (0.0282)}          & \textbf{0.8487 (0.0128)} & \textbf{0.8708 (0.0152)}  \\ \hline
\parbox[t]{2mm}{\multirow{6}{*}{\rotatebox[origin=c]{90}{\textbf{F1-Score}}}}                  & 1                   & 0.8876                   & 0.7561                   & 0.8782                    & 0.8772                            & 0.8637                   & 0.8864                    \\
                                                    & 2                   & 0.8912                   & 0.7902                   & 0.8601                    & 0.8421                            & 0.8572                   & 0.8836                    \\
                                                    & 3                   & 0.8755                   & 0.7537                   & 0.8343                    & 0.8286                            & 0.8477                   & 0.8678                    \\
                                                    & 4                   & 0.8579                   & 0.7542                   & 0.8037                    & 0.8078                            & 0.8332                   & 0.8430                    \\
                                                    & 5                   & 0.8723                   & 0.7989                   & 0.8181                    & 0.8104                            & 0.8529                   & 0.8673                    \\
                                                    & \textbf{Avg. (Std)} & \textbf{0.8769 (0.0133)} & \textbf{0.7706 (0.0221)} & \textbf{0.83888 (0.0304)} & \textbf{0.8332 (0.0283)}          & \textbf{0.8509 (0.0115)} & \textbf{0.8696 (0.0173)}  \\ \hline
\parbox[t]{2mm}{\multirow{6}{*}{\rotatebox[origin=c]{90}{\textbf{TPP}}}}      & 1                   & 0.8394                   & 0.8905                   & 0.8751                    & 0.8333                            & 0.8200                   & 0.8467                    \\
                                                    & 2                   & 0.8662                   & 0.8842                   & 0.8385                    & 0.8652                            & 0.8410                   & 0.8710                    \\
                                                    & 3                   & 0.8575                   & 0.8835                   & 0.8580                    & 0.8725                            & 0.8680                   & 0.8605                    \\
                                                    & 4                   & 0.8575                   & 0.8780                   & 0.8403                    & 0.8590                            & 0.8601                   & 0.8680                    \\
                                                    & 5                   & 0.8409                   & 0.8804                   & 0.8820                    & 0.8698                            & 0.8679                   & 0.8567                    \\
                                                    & \textbf{Avg. (Std)} & \textbf{0.8523 (0.0117)} & \textbf{0.8833 (0.0047)} & \textbf{0.8588 (0.0197)}  & \textbf{0.8600 (0.0158)}          & \textbf{0.8514 (0.0207)} & \textbf{0.8606 (0.0096)}  \\ \hline
\parbox[t]{2mm}{\multirow{6}{*}{\rotatebox[origin=c]{90}{\textbf{FPP}}}}     & 1                   & 0.5749                   & 0.7340                   & 0.7164                    & 0.6988                            & 0.6821                   & 0.7007                    \\
                                                    & 2                   & 0.4225                   & 0.6524                   & 0.6322                    & 0.7002                            & 0.6788                   & 0.7007                    \\
                                                    & 3                   & 0.5661                   & 0.6990                   & 0.6509                    & 0.6880                            & 0.6988                   & 0.6978                    \\
                                                    & 4                   & 0.5966                   & 0.6845                   & 0.6989                    & 0.6782                            & 0.6637                   & 0.6702                    \\
                                                    & 5                   & 0.4256                   & 0.5858                   & 0.6124                    & 0.6501                            & 0.7070                   & 0.5989                    \\
                                                    & \textbf{Avg. (Std)} & \textbf{0.5171 (0.0857)} & \textbf{0.6711 (0.0560)} & \textbf{0.6622 (0.0441)}  & \textbf{0.6831 (0.0205)}          & \textbf{0.6861 (0.0171)} & \textbf{0.6737 (0.0437)}  \\ \hline
\parbox[t]{2mm}{\multirow{6}{*}{\rotatebox[origin=c]{90}{\textbf{PPV}}}} & 1                   & 0.8391                   & 0.7852                   & 0.8092                    & 0.8794                            & 0.8152                   & 0.8854                    \\
                                                    & 2                   & 0.8530                   & 0.7912                   & 0.7796                    & 0.8802                            & 0.8413                   & 0.8732                    \\
                                                    & 3                   & 0.8224                   & 0.7968                   & 0.8255                    & 0.8699                            & 0.8237                   & 0.8779                    \\
                                                    & 4                   & 0.7935                   & 0.8045                   & 0.8103                    & 0.8760                            & 0.8381                   & 0.8637                    \\
                                                    & 5                   & 0.7841                   & 0.7860                   & 0.7998                    & 0.8809                            & 0.8252                   & 0.8802                    \\
                                                    & \textbf{Avg. (Std)} & \textbf{0.8184 (0.0293)} & \textbf{0.7927 (0.0081)} & \textbf{0.8049 (0.0169)}  & \textbf{0.8773 (0.0045)}          & \textbf{0.8287 (0.0108)} & \textbf{0.8761 (0.0082)}  \\ \hline
\parbox[t]{2mm}{\multirow{6}{*}{\rotatebox[origin=c]{90}{\textbf{NPV}}}} & 1                   & 0.9096                   & 0.8040                   & 0.9062                    & 0.9401                            & 0.8843                   & 0.9315                    \\
                                                    & 2                   & 0.9131                   & 0.8838                   & 0.8976                    & 0.9327                            & 0.8909                   & 0.9345                    \\
                                                    & 3                   & 0.9038                   & 0.8224                   & 0.9124                    & 0.9234                            & 0.8790                   & 0.9238                    \\
                                                    & 4                   & 0.8837                   & 0.8108                   & 0.8734                    & 0.9033                            & 0.8648                   & 0.9036                    \\
                                                    & 5                   & 0.9219                   & 0.8638                   & 0.9001                    & 0.9315                            & 0.8814                   & 0.9288                    \\
                                                    & \textbf{Avg. (Std)} & \textbf{0.9064 (0.0143)} & \textbf{0.8370 (0.0350)} & \textbf{0.8979 (0.0149)}  & \textbf{0.9262 (0.0141)}          & \textbf{0.8801 (0.0096)} & \textbf{0.9244 (0.0123)}  \\ \hline
\end{tabular}
\end{table*}

\subsection{Comparative Analysis}\label{sec:results_comparison}
To validate Fair-FLIP, we conduct a five-fold cross-validation against a baseline model and state-of-the-art fairness interventions discussed in Section~\ref{sec:sota}. Table~\ref{tab:all_approaches} reports the performance of the various approaches in terms of both predictive performance and fairness metrics. 

As expected, the \emph{Baseline} approach delivers the highest predictive performance, achieving an average accuracy of 0.8731 and an average F1-score of 0.8769 across the five splits. \emph{Fair-FLIP} demonstrates performance closely comparable to that of \emph{Baseline}, with an average accuracy of 0.8708 (a marginal decrease of only 0.0023) and an average F1-score of 0.8696 (0.0073 lower than \emph{Baseline}). Moreover, Fair-FLIP outperforms \emph{BPFA}, which achieves an average accuracy of 0.8487 and an average F1-score of 0.8509, as well as other benchmark approaches, some of which exhibit significantly lower accuracy, such as the \emph{Pre-Processing} technique, which reaches only 0.7648. 

In terms of the fairness metrics, we first notice that \emph{Baseline} demonstrates an acceptable TPP (0.8523), PPV (0.8184) and NPV (0.9064) but suffers largely in terms of FPP (0.5171). The \emph{Pre-Processing} approach, which compensated heavily on accuracy, demonstrates an improved TPP and FPP with respect to \emph{Baseline} but a slightly degraded PPV and NPV. This shows that while \emph{Pre-Processing} approach can noticeably improve fairness metrics, it considerably reduces accuracy by discarding potentially informative samples. Similarly, \emph{In-process}, in which the predictive performance is like-wise degraded (up to 4\%), demonstrates minor improvement (TPP from 0.8523 to 0.8588) and FPP (from 0.5171 to 0.6622) but shows a slight degradation in terms of PPV and NPV. This shows that the \emph{In-Processing} approach also improves parity measures but at the expense of diminishing accuracy and, as we will show later, additional complexity.
\emph{Post-processing: Threshold-based} approach, on the contrary, demonstrates an improvement in terms of all fairness metrics (TPP by 0.01, FPP by 0.17, PPV by 0.02 and NPV 0.02) at the cost of only 4\% decline in predictive performance in comparison to the \emph{Baseline}. Similarly, \emph{BPFA} effectively address some subgroup bias (FPP increased by 0.17 and PPV by 0.01) but still incur moderate decreases in classification metrics (by 3\%). In contrast, \emph{Fair-FLIP}, which preserves to a large extent the classifier's performance (drop by 0.002), while demonstrating state-of-the-art improvement across all fairness metrics (TPP by 0.01, FPP by 0.16, PPV by 0.06 and NPV 0.02). These results show the effectiveness of Fair-FLIP in mitigating bias (over 10\% in relative numbers) while incurring negligible impact on predictive performance(less than $0.25\%$).

\subsection{Algorithms Complexity}

We now discuss the computational complexity of the various approaches using \emph{big-O} notation\footnote{Due to the different nature and principles of the fairness de-biasing methods, a comparison of execution time is not fair.}.
\begin{table}[ht]
\caption{Computational complexity of the considered fairness methods}
\label{tab:big-o}
  \begin{tabularx}{\linewidth}{ l | X}
\textbf{Method}         & \textbf{Complexity (Big O)} \\ \hline
In-Process              & $O(e \cdot x/b \cdot a)$              \\
Post-Process: Threshold & $O(x+1000 \cdot a) = O(x+a)$        \\
Post-Process: BPFA      & $O(x+a \cdot n \cdot w)$                  \\
Post-Process: Fair-FLIP & $O(x+a \cdot 2 \cdot w) = O(x+a \cdot w)$  
\end{tabularx}
\end{table}
Table~\ref{tab:big-o} reports each algorithm's complexity, omitting pre-processing since it has no dedicated fairness step. The in-process approach adds a fairness penalty (\emph{fairness\_loss}) to each batch. With $a$ protected groups, $x$ total samples, batch size $b$, and $e$ epochs, the complexity grows multiplicatively. By contrast, Post-Process (Threshold-based) performs one pass to collect statistics and then applies fairness adjustments. Threshold tuning explores around 1000 cut-offs between 0.0 and 1.0, adding a constant factor. Bias pruning computes variance scores for each of $n$ neurons and their $w$ weights across $a$ groups, leading to a higher multiplicative term. Fair‐FLIP focuses solely on the final layer's weights. We measure how each feature's variance differs across $a$ protected groups and then re‐scale the weights $w$ of each neuron. For a binary classifier like a deepfake detector, we have two output neurons, thus the Fair-FLIP offers simplified complexity in comparison to BPFA.

\section{Further Analysis and Limitations}\label{sec:discussion}

\begin{figure*}[t]
    \centering
    \includegraphics[width=0.8\linewidth,keepaspectratio]{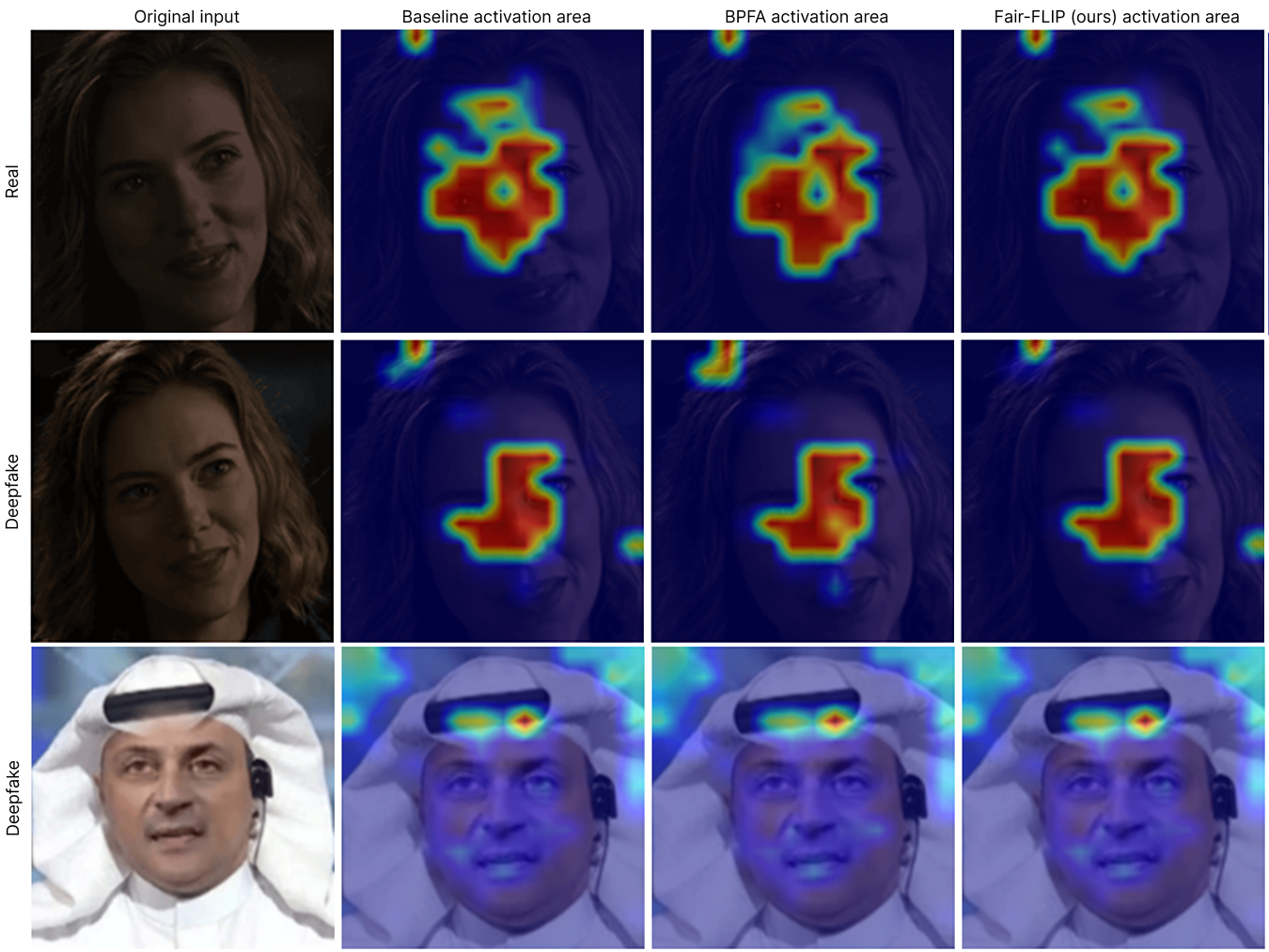}
    \caption{Attention heatmaps for the baseline, BPFA, and Fair-FLIP. In the first row (genuine image), baseline and Fair-FLIP match closely, whereas BPFA's activations are more blurred near the right eyebrow. In the second row, BPFA shows slightly stronger activation in the top‐left and an extra highlight around the mouth, with some missing heat signature on the right; Fair-FLIP's region of interest remains only marginally narrower than the baseline's. In the final example, there are no visible differences between activations, implying no changes in how the model perceives the image.}
    \label{fig:xai}
\end{figure*}

\textbf{Impact on Explainability}. We extend our analysis to investigate the impact of our approach on the model's explainability. We apply \emph{Activation Rollout}~\cite{abnar2020quantifying} on three variants of our deepfake detector (i.e., \emph{Baseline}, \emph{BPFA}, and our proposed approach, Fair-FLIP) and compare the resulting heatmaps. Fig.~\ref{fig:xai} shows an example of resulting heatmaps on one real and two manipulated images. 

Our observations indicate that while both Fair-FLIP and BPFA attention heatmaps are very close to that of the Baseline, the BPFA's aggressive weight pruning may alter the attention patterns more drastically, potentially discarding certain intermediate features. On the contrary, Fair-FLIP's final-layer reweighting is comparatively less invasive, preserving most of the model's discriminative cues. While further analysis is still necessary to draw conclusions, these preliminary investigations show that Fair-FLIP demonstrates an attention similar to that of the Baseline approach. 

\textbf{Ethical Considerations}. Since Fair-FLIP mitigates bias without exposing sensitive user data, i.e., it does not require access to protected attributes or deduce them during inference, it can be considered demographically unintrusive. This characteristic allows us to uphold key ethical principles by promoting fairness without compromising user privacy. This is crucial in real-world applications where demographic profiling raises legal and ethical concerns, such as re-identification risks or unintended discrimination. Moreover, since Fair-FLIP applies adjustments to the final layer rather than intervening in the training process, it maintains transparency and interpretability more than other approaches, such as in-processing techniques.


\textbf{Limitations and Future Work.} The limitations and future works can be summarised as follows. 
\begin{itemize}[noitemsep,leftmargin=*,topsep=0pt]
    \item We demonstrated improvements with respect to a single protected attribute; similarly, unfair discrepancies may be seen simultaneously for multiple attributes. It is valuable to research how Fair-FLIP can counter multi-attribute bias.
    \item The method was tested on a single dataset with annotated demographic attributes. Real-world scenarios often involve more diverse data, including non-human objects or broader environmental contexts.
    \item Further research involves generalising Fair-FLIP to multi-modal settings, like video and audio manipulations, which involve sequential data.
\end{itemize}

\section{Conclusion}
\label{sec:conclusions}
We address the problem of fair deepfake detection. We propose a post-processing bias mitigation technique, Fair-FLIP~(\emph{Fairness-Oriented Final Layer Input Prioritising}), to enhance the fairness of deepfake detection models. Fair-FLIP is based on identifying activations that may introduce demographic bias and then altering final-layer weights based on their suspected demographic bias.
Experimental results comparing Fair-FLIP to baseline approach and bias-mitigation benchmark methodologies show that Fair-FLIP noticeably de-biases the model, enhancing fairness metrics by up to 30\%, while maintaining baseline predictive performance (reduction of 0.25\% in accuracy).

\bibliographystyle{IEEEtran} 

\bibliography{9_references}

\vspace{12pt}

\end{document}